\documentclass[11pt]{article}
\usepackage{acl}
\usepackage{times}
\usepackage{latexsym}
\usepackage[T1]{fontenc}
 \usepackage{amssymb} 
\usepackage[utf8]{inputenc}
\usepackage{microtype}
\usepackage{inconsolata}
\usepackage{graphicx}
\usepackage{booktabs}
\usepackage{multirow}
\usepackage{amsmath}
\usepackage{xcolor}
\usepackage{tikz}
\usepackage{algorithm}
\usepackage{algpseudocode}
\usetikzlibrary{shapes.geometric, arrows, arrows.meta, positioning, fit, backgrounds}
\newcommand\blfootnote[1]{%
  \begingroup\renewcommand\thefootnote{}\footnote{#1}%
  \addtocounter{footnote}{-1}\endgroup}
\newcommand{\srch}{\texttt{<s>}}   
\newcommand{\ans}{\texttt{<a>}}    
\title{StepGap: A Hybrid NLI-LLM Checker for\\
Step-Level Evidence-Gap Detection\\
in Multi-Hop Question Answering}
\author{Yuelyu Ji \\
  School of Computing and Information\\
  University of Pittsburgh \\
  \texttt{yueluji@gmail.com} \\\And
  Zhuochun Li \\
  School of Computing and Information  \\
  University of Pittsburgh \\ \And
  Daqing He \\
  School of Computing and Information  \\
  University of Pittsburgh \\
  \texttt{dah44@pitt.edu} \\ \And}

\author{
 \textbf{Yuelyu Ji\textsuperscript{1}},
  \textbf{Zhuochun Li\textsuperscript{1}},
 \textbf{Hui Ji\textsuperscript{1}},
 \textbf{Daqing He\textsuperscript{1}}
\\
\\
 \textsuperscript{1}School of Computing and Information, University of Pittsburgh
\\
\\
}

\begin{document}
\maketitle
\blfootnote{Code, prompts, and diagnostics available at: 
\url{https://anonymous.4open.science/r/step-checker-0925/}}

\begin{abstract}
We present \textbf{StepGap}, a hybrid NLI-LLM decision tree that detects step-level evidence gaps in multi-hop QA and emits one of three typed labels: \textsc{Contradicted Claim} (CC), \textsc{Irrelevant Evidence} (IE), or \textsc{Missing Bridge} (MB), each tied to a concrete repair action. On 82 multi-hop questions (181 annotated steps, $\kappa{=}0.704$), StepGap reaches sF1$=$72.0, within the bootstrap confidence interval of an LLM-only baseline (70.1) but with a more decomposable structure: every StepGap stage \emph{hurts} F1 when removed, while three of four LLM-only removals \emph{improve} F1---a sign of \emph{competing-error cancellation}, where internal stages mask each other's errors. We further expose a \emph{Q-F1 trap}: question-level F1 is mechanically inflated by checkers that flag every step, making step-level F1 the necessary diagnostic. Used as a typed GRPO process reward, StepGap improves Qwen2.5-7B-Instruct Exact Match from $32.1{\pm}0.3$ to $35.4{\pm}0.9$ across three seeds, with the single-run comparison showing a $+5.6$ Avg EM gain over the matched Search-R1 GRPO reproduction.

\end{abstract}

\begin{figure}[t]
\centering
\includegraphics[width=\columnwidth
]{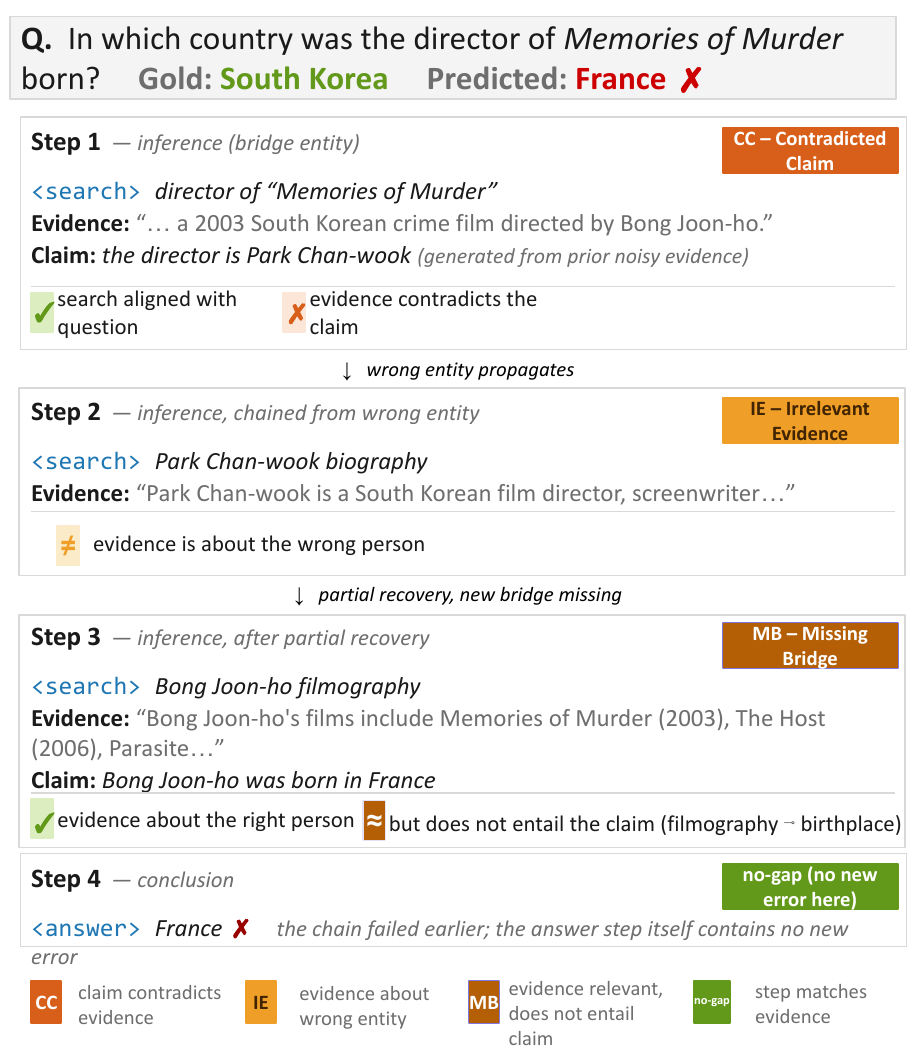}
\caption{StepGap on a synthetic 4-step trace. Endpoint EM only sees the wrong final answer; StepGap typed-localizes CC, IE, and MB at the steps where each gap first emerges, with each gap type mapped to a distinct repair action.}
\label{fig:case_study}
\end{figure}

\section{Introduction}

Multi-hop question answering (QA) requires retrieving, connecting, and verifying evidence across several intermediate steps \cite{yang2018hotpotqa,ho2020constructing,trivedi2022musique}. Yet both evaluation and reinforcement-learning training in search-interleaved frameworks remain endpoint-oriented: systems are scored and rewarded on final-answer Exact Match (EM) alone.

Endpoint-only signals have two structural defects. \emph{First}, EM cannot distinguish an evidence-grounded chain from a spuriously correct one: a policy can reach the right answer through shortcuts that bypass retrieval or fill missing facts from parametric memory~\cite{jin2025search,min2023factscore}. 
\emph{Second}, EM provides no localization on wrong answers: the policy receives zero reward with no signal about \emph{which} step broke, so it cannot learn step-specific repair behavior~\cite{lightman2023lets,lu2025doctorrag}.
We call such localized step-level failures \emph{evidence gaps}, and distinguish three gap types. A \textsc{Contradicted Claim} (CC) occurs when a step asserts a claim that conflicts with the retrieved evidence; \textsc{Irrelevant Evidence} (IE) occurs when the evidence does not address the entity or relation needed by the step; and \textsc{Missing Bridge} (MB) occurs when the evidence is relevant but does not yet entail the step's claim. While step-level error taxonomies have been explored before~\cite{deshpande2505trail,lu2025doctorrag}, our \textsc{CC/IE/MB} types are explicitly designed for \emph{repair-action coupling}: each maps to a single concrete operation (retract / re-search / bridging search) that an RL policy can be trained to perform. 
Figure~\ref{fig:case_study} illustrates this: three structurally distinct gap types can cascade into a single undifferentiated EM$=$0---a contradicted claim requiring retraction, irrelevant evidence requiring re-search, and a missing bridge requiring an additional bridging search.

Existing step-level checkers either collapse all subtasks into a single LLM call \cite{xiong2025rag,wang-etal-2024-math} or rely on a single NLI call \cite{tang2024minicheck,laurer2024deberta}, neither of which simultaneously emits typed labels and grounds them in deterministic decisions. Post-hoc and claim-level verifiers \cite{gao2023rarr,min2023factscore,ru2024ragchecker} operate after a complete answer and provide no training-time
signal at all.
\textbf{StepGap} is a five-stage hybrid decision tree: 
three LLM stages handle open-ended reading subtasks 
(alignment, abstention, entity consistency and quote 
extraction), and two NLI stages handle deterministic 
entailment (local and cross-step). The tree terminates at 
a typed gap label (\textsc{CC}/\textsc{IE}/\textsc{MB}) 
coupled to a concrete repair action, and the typed output 
enters GRPO as a per-step process reward, training the 
policy to perform the prescribed repair when a gap fires.
\paragraph{Our contribution.}
We make three claims. 
\textbf{First}, typed step-level gap labels are useful only when
they are coupled to repair actions: \textsc{CC} maps to
retraction, \textsc{IE} to re-search, and \textsc{MB} to
bridging search.

\textbf{Second}, StepGap is more decomposable than an
LLM-only checker. On the 181-step human benchmark,
StepGap reaches sF1$=$72.0, within the bootstrap confidence
interval of the LLM-only baseline (70.1), but every StepGap
component hurts F1 when removed. In contrast, three of four
LLM-only removals improve F1, revealing
\emph{competing-error cancellation}. We also expose a
\emph{Q-F1 trap}, showing why question-level detection must
be paired with step-level F1 and balanced accuracy.

\textbf{Third}, the typed signal works as a process reward.
A Typed+Shape GRPO reward built on StepGap improves
Qwen2.5-7B EM from $32.1{\pm}0.3$ to $35.4{\pm}0.9$
across three seeds and raises answer-claim support to
$71.5\%$; a distilled Qwen2.5-7B student retains $87\%$
of teacher F1 with no API calls.



\begin{figure*}[t]
\centering
\includegraphics[width=0.95\textwidth]{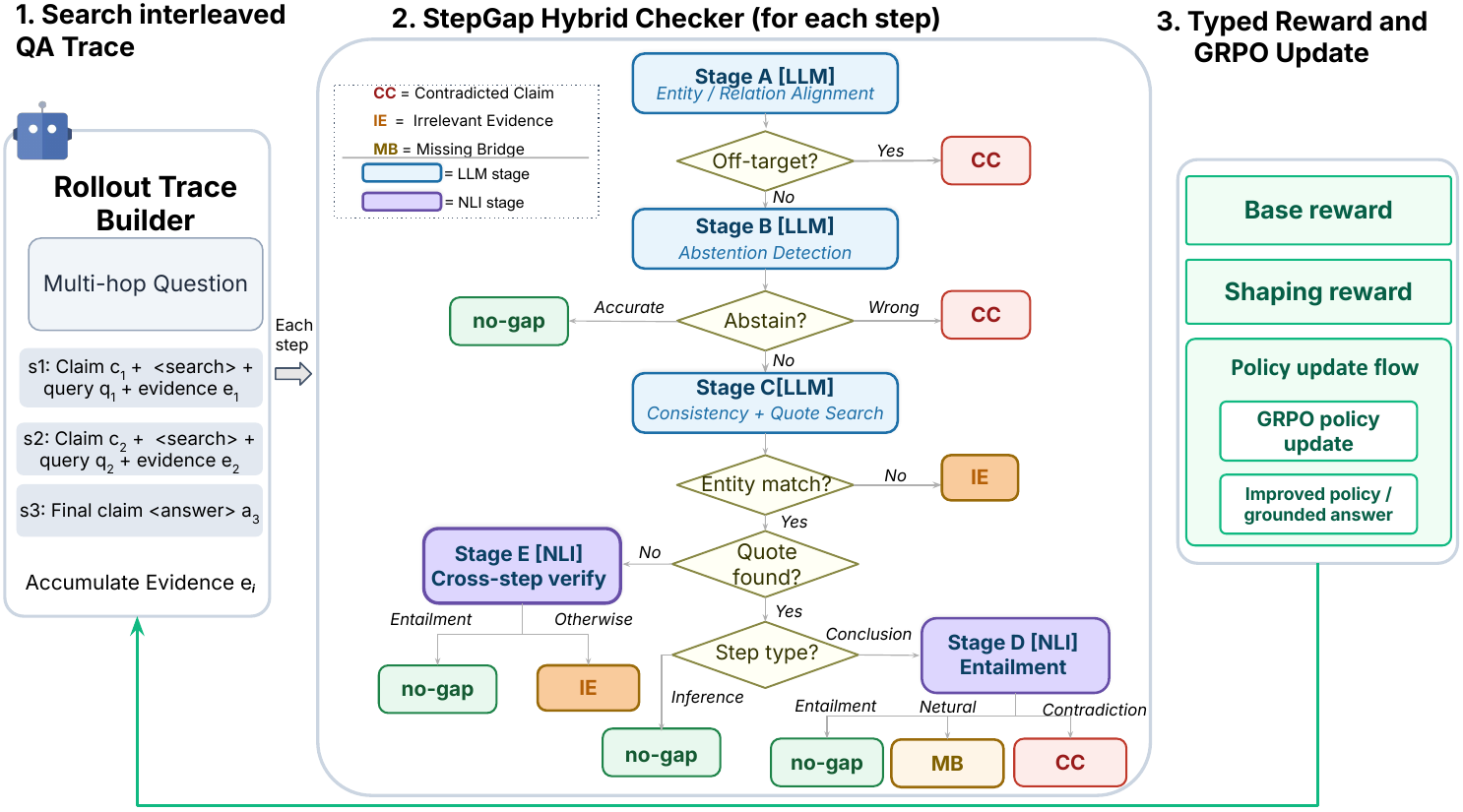}
\caption{StepGap pipeline.
\textbf{(1)} The generator produces a search-interleaved
trace; each step $s_i=(c_i,q_i,e_i,a_i)$ is fed to the
checker with accumulated evidence $e_{\leq i}$.
\textbf{(2)} The hybrid checker traverses five typed stages
(LLM: A~alignment, B~abstention, C~entity+quote;
NLI: D~local entailment, E~cross-step verification)
and emits $\tau_i\in\{\text{no-gap},
\textsc{CC},\textsc{IE},\textsc{MB}\}$.
\textbf{(3)} $\tau_i$ drives a typed GRPO reward
($r^{\text{base}}+r^{\text{shape}}$, \S\ref{sec:reward}),
closing the loop back to the generator.
\textsc{CC}~=~Contradicted Claim;
\textsc{IE}~=~Irrelevant Evidence;
\textsc{MB}~=~Missing Bridge.}

\label{fig:pipeline}
\end{figure*}

\section{Related Work}
\paragraph{Search-interleaved reasoning.}
IRCoT~\cite{trivedi2023interleaving},
ReAct~\cite{react2023}, Search-R1~\cite{jin2025search}, and
R1-Searcher~\cite{song2025r1} train search policies with
final-answer EM rewards and cannot attribute failure to
specific steps; on a wrong answer, EM-only reward provides no
gradient about which step broke. StepGap supplies a per-step
typed signal during rollout, enabling the generator to learn
gap-specific repair behaviors rather than a uniform
``avoid errors'' objective.

\paragraph{Post-hoc verification.}
RARR~\cite{gao2023rarr}, CRITIC~\cite{gou2023critic}, and
Chain-of-Verification~\cite{dhuliawala2023cove} apply
verification \emph{after} a complete answer is produced, so
they cannot localize mid-chain failures or provide
training-time signal. StepGap is invoked inline at each step
during rollout, redirecting the generator before the chain
propagates a failure.

\paragraph{Claim-level faithfulness.}
FActScore~\cite{min2023factscore},
MiniCheck~\cite{tang2024minicheck},
SelfCheckGPT~\cite{manakul2023selfcheckgpt},
RAGAS~\cite{es2023ragas}, and
RAGChecker~\cite{ru2024ragchecker} score whether generated
output is supported by retrieved evidence, but do not capture
reasoning-chain structure (e.g., whether an intermediate query
targeted the right entity); a model trained on output-level
faithfulness can still reach correct answers via ungrounded
intermediate steps. StepGap evaluates each step against its
own evidence, catching failures where they first occur.

\paragraph{Process rewards and agentic RAG.}
Process Reward Models~\cite{lightman2023lets,wang-etal-2024-math}
provide step-wise scalar supervision, with ReARTeR
\cite{sun2025rearter} adding MCTS, and RAG-Gym
\cite{xiong2025rag} and ProRAG using LLM-judge process rewards.
TRAIL~\cite{deshpande2505trail} introduces a trace-level
typed error ontology, and Doctor-RAG~\cite{lu2025doctorrag}
localizes the earliest failing step for post-hoc repair. We extend this line on
three axes simultaneously: \emph{step granularity}, \emph{NLI-anchored typed labels}, and \emph{repair-action coupling}. This is a conjunction
that is jointly necessary for a typed behavioral reward but
has not previously been combined in a single system.

\paragraph{Hybrid verifiers.}
NLI--LLM hybrids have been explored for output-level
factuality~\cite{xiong2025toward,liu2025examining,tang2024minicheck,ji2025mrag}
but not as process-reward signals. Scoring a final answer
and scoring an intermediate step are structurally different;
we show that an LLM-only checker, even a strong one, collapses
to near-uniform IE labels when applied per-step
(Appendix~\ref{app:lens2_dist}), motivating the hybrid design.
To our knowledge, prior work has not jointly demonstrated: (i)~hybrid
checkers exhibit a healthier ablation profile than LLM-only
variants for step-level gap detection, and (ii)~such checkers
can serve as typed process rewards for search-interleaved RL
training.

\section{Problem Formulation}

\paragraph{Reasoning traces.}
We model a search-interleaved multi-hop QA trace as an ordered
sequence $\mathcal{T} = \{s_1, \ldots, s_n\}$. Each
\textbf{step} $s_i$ is a tuple
$s_i = (c_i, q_i, e_i, a_i)$, where
$c_i$ is the model's intermediate reasoning text (a natural-language claim or sub-conclusion produced before the action),
$q_i$ is an optional \texttt{<search>} query,
$e_i$ is the evidence retrieved by $q_i$, and
$a_i$ is an optional \texttt{<answer>} block emitting the final
answer $\hat{a}$. We call $s_i$ an \emph{inference step} when $q_i \neq \varnothing$ and $a_i = \varnothing$ (the model issues a search and writes a sub-claim), and a \emph{conclusion step} when $a_i \neq \varnothing$ (the model commits to the final answer). The checker takes the entire tuple plus all earlier evidence as input, so its judgment depends jointly on the intermediate claim, the search query, the retrieved snippets, and the answer if present.

Throughout, \textbf{step} refers to a unit of the reasoning trace ($s_i$); \textbf{stage} refers to a node in the checker's decision tree (Stages~\textsc{A}--\textsc{E}, indexed by letters to visually separate them from $s_i$). One checker invocation traverses up to five stages on a single reasoning step.

\paragraph{Step-level evidence gap detection.}
Step $s_i$ is \emph{gap-free} if its claim---$c_i$ when  $s_i$ is an inference step, $a_i$ when $s_i$ is a conclusion step---is logically entailed by the accumulated evidence $\bigcup_{j \leq i} e_j$.
Otherwise, the step has a \emph{gap}, characterized by gap type $\tau \in \{\textsc{CC}, \textsc{IE}, \textsc{MB}\}$. The checker $\mathcal{C}$ maps each $(s_i, e_{\leq i})$ to a structured output $(\tau, \textit{confidence}, \textit{rationale})$. The five stages of $\mathcal{C}$ are described in Section~\ref{sec:checker}.



\section{Hybrid Step-Gap Checker: Decision Tree}
\label{sec:checker}

For each step $s_i = (c_i, q_i, e_i, a_i)$ with accumulated evidence $e_{\leq i}$, the checker traverses up to five stages (Figure~\ref{fig:pipeline}, panel~2). Each stage either emits a typed gap label and stops, or hands off to the next stage; if no stage fires, the step is labeled no-gap. We organize this section by traversal order (Stage~A through Stage~E) so that the tree's structure is visible end-to-end; Table~\ref{tab:taxonomy_signals} summarizes which stages can emit which gap type. Below, \emph{the step's claim} means $c_i$ when $s_i$ is an inference step and $a_i$ when $s_i$ is a conclusion step. The search query $q_i$ is consulted only at Stage~A and is never used as an NLI hypothesis.

\begin{table}[t]
\centering
\small
\caption{Stages of the StepGap decision tree, the gap
type each can emit, and the NLI-grounded typed signal at that
node. Stages~D--E are NLI; A--C are LLM. Each gap type
maps to a single repair action used downstream as the
typed reward signal (\S\ref{sec:reward}).}
\label{tab:taxonomy_signals}
\begin{tabular}{l l l l}
\toprule
\textbf{Stage} & \textbf{Type} & \textbf{Signal}
& \textbf{Repair} \\
\midrule
A (LLM) & CC  & alignment fail        & retract \\
B (LLM) & CC  & wrong abstention      & retract \\
C (LLM) & IE  & entity mismatch       & re-search \\
D (NLI) & CC  & contradiction         & retract \\
D (NLI) & MB  & neutral, quote found  & bridging search \\
E (NLI) & IE  & no entailing prior    & re-search \\
\bottomrule
\end{tabular}
\end{table}

\paragraph{Stage A -- Alignment.}
Stage~A checks whether the step targets the correct entity 
and relation, \emph{before} retrieval is consulted. This 
matters because a mis-targeted step cannot be rescued by 
retrieval---any fetched document is irrelevant by 
construction, and surface-token overlap can masquerade as 
support. We classify mis-targeting into three drift 
patterns---relation drift (wrong predicate), entity drift 
(wrong subject), and scope drift (sub-part vs.\ whole)---
and emit \textsc{CC} on any drift.
\paragraph{Stage B -- Abstention validity.}
Stage~B handles steps that assert ``cannot be determined,'' 
verifying that the abstention is grounded in the evidence. 
An unjustified abstention is functionally a contradicted 
claim: the step commits to ``the answer is unknown'' when 
the evidence rules that out. A grounded abstention exits 
to no-gap; a wrong abstention emits \textsc{CC}.
\paragraph{Stage C -- Entity consistency and quote.}
Stage~C verifies that the retrieved evidence is about the 
targeted entity and, if so, extracts a verbatim supporting 
quote. Without this filter, off-topic documents sharing 
lexical tokens with the claim would pass to NLI as spurious 
premises. If the page's title and lead sentence describe a 
different entity, Stage~C emits \textsc{IE} directly; 
otherwise the checker finds a 5--20 word verbatim span on 
the same entity, relevant to the claim. A missing quote 
routes a conclusion step to Stage~E and an inference step 
to no-gap (a search query is not a well-formed NLI 
hypothesis).
\paragraph{Stage D -- Local entailment.}
Stage~D decides whether the Stage~C quote actually entails 
the step's claim. Entity match plus a returned quote is 
necessary but not sufficient: the quote can be on-topic yet 
fail to imply the claim (a missing-bridge case) or overtly 
contradict it. A single NLI call with the quote as premise 
and the claim as hypothesis resolves this: 
\textsc{entailment}~$\to$~no-gap; 
\textsc{neutral}~$\to$~\textsc{MB} 
(\emph{repair: bridging search}); 
\textsc{contradiction}~$\to$~\textsc{CC}.
\paragraph{Stage E -- Cross-step verification.}
Stage~E handles conclusion steps where Stage~C found no 
local quote, asking whether the claim is entailed by 
evidence retrieved at earlier inference steps. Multi-hop 
answers typically rest on facts established several steps 
earlier, so the effective evidence pool is the union of 
all prior retrievals. Stage~E iterates over entity-matched 
prior snippets as candidate premises: an entailing snippet 
exits to no-gap; otherwise the step emits \textsc{IE} 
(\emph{repair: re-search}). The entity-match constraint 
prevents off-topic earlier retrievals from producing 
spurious entailments.
\section{Using the Typed Signal as a Process Reward}
\label{sec:reward}

A checker that only \emph{detects} gaps does not change policy behavior: the model can detect a gap at step~$i$ and still commit at step~$i{+}1$. We inject the typed output into RL training as a per-step process reward so the policy is reinforced to execute the prescribed repair when a gap fires. The reward has two parts: a base scalar conditioned on gap type (\S\ref{sec:base_reward}) and an action-conditional shaping term that verifies the repair is performed (\S\ref{sec:shape_reward}); both plug into GRPO unmodified (\S\ref{sec:reward_grpo}).

\subsection{Base Reward by Gap Type}
\label{sec:base_reward}

Let $\tau_i \in \{\text{no-gap}, \text{MB}, \text{IE}, \text{CC}\}$
be the gap type emitted by the checker for step $s_i$. The base
reward is
\begin{equation}
r^{\text{base}}(\tau_i) = \begin{cases}
+0.20 & \text{no-gap} \\
-0.05 & \text{MB} \\
-0.10 & \text{IE} \\
\mathbf{+0.05} & \text{CC detected}
\end{cases}
\end{equation}

Three choices are non-obvious. \emph{No-gap is positive} ($+0.20$), following standard process-reward practice \cite{lightman2023lets,wang-etal-2024-math,xiong2025rag,hiprag2025}. \emph{CC receives a small detection bonus} ($+0.05$): the reward is assigned after the checker detects a contradiction, not to the contradicted claim itself. Penalizing all detected CC labels would instead teach the generator to hide contradictions behind vaguer claims that fewer NLI entailment checks can reject~\cite{jiang2025rag}. The bonus is kept small so it does not compete with EM and only rewards exposing a contradiction that the next-step shaping term can require the policy to retract.. \emph{IE is penalized more than MB} ($-0.10$ vs.\ $-0.05$): IE requires reformulating the query, MB only a bridging search; the asymmetry pushes the generator toward better initial queries.


\subsection{Behavioral Shaping Reward}
\label{sec:shape_reward}
We add a \emph{shaping reward} conditional on the repair action at step $s_i$ given the gap type at step $s_{i-1}$, following the action-conditional design of \textbf{RAG-Gym} \cite{xiong2025rag} and the process-reward formulation of \cite{zhang2025process}.


\begin{equation}
\small
r^{\text{shape}}(\tau_{i-1}, s_i) = \begin{cases}
{+0.10} & \text{IE/MB}_{i-1} \wedge \srch_i \wedge q_i \neq q_{<i} \\
{+0.15} & \text{CC}_{i-1} \wedge \text{retract}_i \\
{-0.15} & \text{IE/MB/CC}_{i-1} \wedge \ans_i \\
{-0.05} & \text{IE/MB}_{i-1} \wedge \srch_i \wedge q_i \approx q_{<i} \\
0       & \text{otherwise}
\end{cases}
\end{equation}

Here $q_i$ denotes the query string of a search step; we consider $q_i \approx q_{<i}$ if its token-F1 against any prior query exceeds 0.7---a threshold selected on the same $200$-iteration pilot used for base reward weights (Appendix~\ref{app:hparams}). Retraction is detected by a small set of surface patterns (``actually'', ``wait'', ``correction'', ``I was wrong''), combined with a check that the new claim differs from the previous one.

The shaping reward $r^{\text{shape}}$ defined above implements the principle ``detection without repair is failure'':  ignoring the checker after a gap costs $-0.15$ (a step-level analogue of Reflexion's self-correction~\cite{shinn2023reflexion}), executing the repair earns $+0.10$ (re-search) or $+0.15$ (retraction; CC's higher shaping bonus reflects the larger behavioral cost of retracting a contradicted claim), and lazy near-duplicate queries earn $-0.05$ to prevent gaming the shaping signal (cf.~\cite{mmsearchr1}).

\subsection{Plugging the Typed Reward into GRPO}
\label{sec:reward_grpo}

The checker enters GRPO~\cite{shao2024deepseekmath} only at the return-computation step. During rollout, $\mathcal{C}$ is invoked once per emitted step $s_i$, returning $\tau_i$, which is cached with the step's token span $[\ell_i, r_i)$. The trajectory return is
\begin{equation}
\label{eq:final_reward}
R = r^{\text{EM}} + \lambda \sum_{i=1}^{n}\!
\left[r^{\text{base}}(\tau_i) + r^{\text{shape}}(\tau_{i-1}, s_i)\right],
\end{equation}
with $r^{\text{EM}} \in \{0, 1\}$, $\lambda = 1.0$, and
$\tau_0 \,{=}\, \text{no-gap}$ at the boundary. We use the
\emph{dense} variant: each step's bracket is assigned only to tokens in $[\ell_i, r_i)$, and the group-relative standardization $\hat{A} = (R - \mu_R)/\sigma_R$ is computed token-wise. The clip ratio, KL penalty, and standardization are kept at GRPO defaults; only the return $R$ in Eq.~\eqref{eq:final_reward} differs across our reward ablations, so any EM gain attributes to the typed checker signal and shaping rather than to optimizer re-tuning.

\section{Checker Evaluation}
\label{sec:lens1_sf1}

\subsection{Human-Annotated Evaluation Set}

We sample 82 questions from HotpotQA, 2WikiMultiHopQA, and
MuSiQue, run Qwen2.5-7B-Instruct to produce full search-
interleaved traces, and manually annotate 181 individual steps
(see Appendix~\ref{app:annotation} for protocol). Each step is
labeled as no-gap or gap (with type $\in$
$\{$\textsc{CC}, \textsc{IE}, \textsc{MB}$\}$). Two annotators independently label the test subset. 
Inter-annotator agreement is Cohen's $\kappa = 0.704$, which we interpret as moderate agreement per \citet{mchugh2012interrater}, consistent with the inherently subjective nature of distinguishing \textsc{MB} from \textsc{IE} near the decision boundary. Disagreements were adjudicated by a third senior annotator, and the post-adjudication consensus serves as the ground-truth label set used in all evaluations (full protocol in Appendix~\ref{app:annotation}). 

 Per-type $\kappa$ (reported in Appendix~\ref{app:annotation}) is highest for IE and CC, and lower for MB, where MB's low frequency ($n{=}9$) inflates sampling variance. Table~\ref{tab:benchmark_stats} reports dataset composition.

\paragraph{Silver-standard scale check.}
To complement the $181$-step gold evaluation set, we measure StepGap's agreement with the GPT-4.1-mini+NLI teacher checker on the $1{,}000$-step distillation pool (Appendix~\ref{app:distillation}, disjoint from the human benchmark). 
Treating the teacher checker's labels as a \emph{silver-standard} reference---i.e., a high-quality but model-generated label set that substitutes for human gold labels when annotation scale is impractical---StepGap reaches $\mathrm{F1}{=}0.88$ on the binary \texttt{has\_gap} dimension and $\mathrm{F1}{=}0.81$ on typed CC/IE/MB. This high agreement at $\sim 5\times$ the gold-evaluation-set scale indicates that StepGap's behavior on the 181-step human evaluation set is representative of its behavior at scale, not a sampling artifact.

\begin{table}[t]
\centering
\small
\caption{Human-annotated evaluation set 
statistics. Gap\% = fraction of steps labeled 
as gap; Acc\% = question-level accuracy.
GRPO training uses 169K NQ+HotpotQA questions
(\texttt{PeterJinGo/nq\_hotpotqa\_train});
standard dev splits: HotpotQA 7,405~Q,
2Wiki 12,576~Q, MuSiQue 2,417~Q
(Appendix~\ref{app:exp_stdsplits}).}
\label{tab:benchmark_stats}
\begin{tabular}{lrrrr}
\toprule
\textbf{Dataset} & \textbf{\#Q} 
  & \textbf{\#Steps} & \textbf{Gap\%} 
  & \textbf{Acc\%} \\
\midrule
2WikiMultiHopQA  & 29  & 66  & 62 & 21 \\
HotpotQA         & 23  & 53  & 57 & 17 \\
MuSiQue          & 30  & 62  & 58 & 10 \\
\midrule
\textbf{Total}   & \textbf{82} 
  & \textbf{181} & \textbf{59} 
  & \textbf{16} \\
\bottomrule
\end{tabular}
\end{table}

We compare StepGap against four internal variants
\textbf{LLM-Strict} and \textbf{LLM-Only}: GPT-4.1-mini at every stage, differing only in the gap-flagging threshold (\textbf{LLM-Strict}: any single stage-level gap signal suffices; \textbf{LLM-Only}: requires the checker's overall confidence $\geq 0.5$); \textbf{Single-LLM-XL}: GPT-5-mini at every stage, no NLI; \textbf{NER-Quote}: spaCy NER replacing Stage~C's LLM quote search) and two external baselines (\textbf{MiniCheck} \cite{tang2024minicheck}; \textbf{NLI-only} \cite{laurer2024deberta}: StepGap's DeBERTa-v3 applied directly to (full evidence, claim) without LLM stages). 
We also considered answer-level faithfulness checkers (RAGAS~\cite{es2023ragas}, FActScore~\cite{min2023factscore}, RAGChecker~\cite{ru2024ragchecker}), but these score \emph{complete answers} rather than intermediate steps and are therefore not directly comparable as step-level checkers; among them MiniCheck is the closest step-applicable representative and is included as a baseline.
Table~\ref{tab:variants_sf1} reports performance with bootstrap $95\%$ CIs. We report both step-level F1 (sF1: per step, does \texttt{has\_gap} match the human label?) and question-level F1 (Q-F1: on each wrong-answer question, did
the checker detect any gap?), sF1 paired with balanced accuracy (BA) is the more diagnostic of the two: Q-F1 fires whenever \emph{any} gap is detected on a wrong-answer question, so a checker that flags every step trivially achieves near-ceiling Q-F1 without distinguishing real gaps from spurious ones. sF1, in contrast, penalizes per-step false positives, and BA further corrects for the $\sim 60\%$ gap-rate class imbalance---together they capture both detection coverage and precision.
\begin{table*}[t]
\centering
\small
\caption{Step-level performance on the 181-step evaluation set.
sF1/sP/sR: step-level F1, precision, recall.
Q-F1: a question is TP iff the checker detects $\ge 1$
gap on a wrong-answer chain. BA: balanced accuracy.
$\dagger$~degenerate variants whose Q-F1 is mechanically
inflated by flagging nearly every step as IE
(\S\ref{sec:lens2_dist}). $\ddagger$~binary checkers
without typed CC/IE/MB output. Brackets: 2000-iter
percentile bootstrap 95\% CIs.}

\label{tab:variants_sf1}
\begin{tabular}{lccccc l}
\toprule
\textbf{Variant} & \textbf{sF1 [95\% CI]} & \textbf{sP} & \textbf{sR}
                 & \textbf{Q-F1} & \textbf{BA}
                 & \textbf{Note} \\
\midrule
\multicolumn{7}{l}{\emph{Internal variants (full typed CC/IE/MB output)}} \\
LLM-Strict            & 67.2 [60.4, 73.8]   & 60.3 & 76.0
                  & 89.2  & 47.8
                  & high FP, recall-heavy \\
LLM-Only              & 70.1 [63.5, 76.2]   & 69.2 & 71.2
                  & 81.8  & 44.4
                  & competing-error baseline \\
\textbf{StepGap (ours)} & \textbf{72.0} [65.6, 77.8]
                  & \textbf{71.0} & 73.1
                  & 81.2  & 60.6
                  & best P/R balance, healthy ablation \\
\midrule
\multicolumn{7}{l}{\emph{Degenerate alternatives (full output but collapsed onto IE)}} \\
Single-LLM-XL         & 69.2 [63.0, 75.1]   & 57.2 & 87.5
                  & 91.4$^\dagger$ & 50.0$^\dagger$
                  & all flagged; IE$=$87.8\%, CC$=$0\% \\
NER-Quote             & 68.0 [61.5, 74.2]   & 56.8 & 84.6
                  & 91.4$^\dagger$ & 50.0$^\dagger$
                  & NLI fires on noun phrases \\
\midrule
\multicolumn{7}{l}{\emph{Binary-only baselines (no typed gap output)$^\ddagger$}} \\
NLI-only \cite{laurer2024deberta}
                  & 61.7 [54.3, 68.2]   & 63.3 & 53.5
                  & 75.2 & 49.7
                  & no LLM stages; pure entailment \\
MiniCheck \cite{tang2024minicheck}
                  & 67.3 [58.3, 73.2]   & 61.2 & 72.1
                  & 90.4 & 51.5
                  & external published checker \\
\bottomrule
\end{tabular}
\end{table*}

\paragraph{Statistical interpretation.}
\emph{(1)} StepGap and LLM-Only are statistically indistinguishable on top-line sF1 (CIs overlap at $n{=}181$). 
Our argument for StepGap instead rests on \emph{compositional} ablation properties (Appendix~\ref{app:hybrid_rationale}): every StepGap component removal \emph{hurts} F1, meaning each stage contributes real signal that the rest of the system cannot replace---the checker is \emph{decomposable}. By contrast, three of four LLM-only component removals \emph{improve} F1, meaning those components were actively introducing errors that other components had been masking; this is \emph{competing-error cancellation}, and the checker is not decomposable.
\emph{(2)} Single-LLM-XL and NER-Quote expose what we call a \textbf{Q-F1 trap}: Q-F1 fires whenever \emph{any} step is flagged on a wrong-answer question, so a checker flagging every step achieves near-ceiling Q-F1 mechanically. Both variants flag ${\sim}95\%$ of steps and reach Q-F1$=$91.4 while sF1 collapses to $\leq 69$ and BA to chance ($50\%$). A flag-everything checker reaches Q-F1$=2w/(1+w)$ in closed form, giving $0.67/0.78/0.91$ at $w{=}0.5/0.65/0.84$; a dummy reproduces this within $\pm 0.02$ on four benchmarks (Appendix~\ref{app:qf1_trap_empirical}). Q-F1 alone cannot distinguish a precise checker from one that always fires.
\emph{(3)} 
The three metrics must be read jointly: 
Q-F1 is inflated by flag-everything (point 2), sF1 can 
mask class imbalance (a no-gap predictor scores 
$\sim 60$ at our $59\%$ gap rate), and BA alone ignores 
precision. StepGap is the only variant competitive on 
all three (sF1$=$72.0, Q-F1$=$81.2, BA$=$60.6\%); 
Single-LLM-XL wins Q-F1 but loses sF1 and BA, falling 
into the trap.


Our claim is not that StepGap outperforms LLM-only on sF1 alone, but that its component ablations are \emph{monotone}: every removal hurts F1. This matters because a system whose components consistently add signal admits modular debugging and distillation (Appendix~\ref{app:distillation}). LLM-only's ablations swing oppositely ($+1.6/+0.8/+1.7$ vs $-6.5$), so stage effects are entangled.

\subsection{Category Distribution as a Health Indicator}
\label{sec:lens2_dist}

The distribution of emitted gap types diagnoses \emph{checker health}, orthogonal to sF1: a checker can score moderately on sF1 yet be useless as a typed reward source if it cannot reliably distinguish CC/IE/MB. LLM-Only and StepGap both produce IE in the $38$--$41\%$ range, consistent with retrieval failure being the dominant mode in annotation. We treat this 38--41\% range as a benchmark-specific sanity check rather than a universal target distribution. The two degenerate variants leave this band entirely. Single-LLM-XL collapses to $87.8\%$ IE with zero CC and zero MB: the strongest single LLM yields a \emph{less} discriminating checker. NER-Quote reaches $74\%$ IE because spaCy noun-phrase queries are not well-formed NLI hypotheses, so Stage~D over-reports retrieval failure. Both pathologies are threshold-independent (Appendix~\ref{app:thresh_sweep}), confirming the collapse is architectural. A confusion-matrix view (Appendix~\ref{app:gap_correlation}) shows StepGap flags a gap on $56/69$ wrong-answer questions and $12/13$ correct-answer questions, exposing the \emph{lucky-correct} regime endpoint EM cannot see; the full first-gap distribution is in Appendix~\ref{app:first_gap_localization}.

\section{GRPO Training with Typed Reward}
\label{sec:lens4_downstream}

\subsection{Setup}
A high-quality checker does not by itself train a policy to fix the gaps it detects. We therefore test whether StepGap's typed outputs work as a per-step process reward, holding backbone, training mix, and hyperparameters constant. Only checkers emitting typed CC/IE/MB qualify: StepGap (ours) and LLM-Only; binary support-checkers (MiniCheck, NLI-only) are excluded by design and remain evaluated on checker quality alone (\S\ref{sec:lens1_sf1}).

We train Qwen2.5-7B-Instruct with GRPO \cite{shao2024deepseekmath} on NQ+HotpotQA (Table~\ref{tab:benchmark_stats}); retrieval uses E5-base-v2 \cite{wang2022text} over Wikipedia 2018 (top-$k{=}5$). The teacher checker (GPT-4.1-mini) is called ${\sim}72$K times per $3{,}000$-iteration run at ${\sim}\$22$; the distilled 7B student (Appendix~\ref{app:distillation}) eliminates this cost. We compare four reward configurations: \textbf{Search-Only} (EM only), \textbf{Binary Gap} ($+0.2$ no-gap / $0$ gap; prior design), \textbf{Typed Base}, and \textbf{Typed+Shape (ours)}.

\subsection{Main Results}
Table~\ref{tab:rl_results} reports per-dataset and Avg EM. The headline result is Typed+Shape's $35.4$ Avg EM, a $+5.6$ gain over our directly comparable Search-R1 GRPO baseline ($29.8$, same backbone and training mix), with the gain holding across all three datasets. Three-seed mean$\pm$std and the Llama-3.1-8B backbone-transfer study are in Appendix~\ref{app:grpo_robustness}; auxiliary grounding metrics in Appendix~\ref{app:rl_grounding}.


\begin{table*}[t]
\centering
\small
\caption{GRPO results on standard dev splits. \textbf{Top}: published baselines comparable to ours (same Qwen2.5-7B-Instruct backbone). \textbf{Middle}: external systems with non-comparable training setup (different base or different training mix). \textbf{Bottom}: our reward ablations on the same NQ+HotpotQA training mix and Qwen2.5-7B-Instruct backbone as Search-R1. The directly comparable row is Search-R1 (GRPO); \textbf{Typed+Shape gains $+5.6$ Avg EM over it ($29.8\to 35.4$)}. Bottom-panel ``Search-Only'' is the single-seed Search-R1 GRPO reproduction; three-seed mean$\pm$std for all our ablations (where Search-Only reaches $32.1{\pm}0.3$) is in Appendix~\ref{app:grpo_robustness}, Table~\ref{tab:grpo_robustness}.
}
\label{tab:rl_results}
\begin{tabular}{l l l c c c c}
\toprule
\textbf{Method} & \textbf{Backbone} & \textbf{Train mix}
                & \textbf{HQA} & \textbf{2Wiki}
                & \textbf{MuSiQue} & \textbf{Avg} \\
\midrule
\multicolumn{7}{l}{\emph{Same backbone (Qwen2.5-7B-Inst), comparable to ours}} \\
Naive RAG          & 7B-Inst & ---       & 29.9 & 23.5 &  5.8 & 19.7 \\
IRCoT              & 7B-Inst & ---       & 13.3 & 14.9 &  7.2 & 11.8 \\
SFT                & 7B-Inst & NQ+HQA    & 21.7 & 25.9 &  6.6 & 18.1 \\
Search-R1 (PPO)    & 7B-Inst & NQ+HQA    & 37.0 & 41.4 & 14.6 & 31.0 \\
Search-R1 (GRPO)   & 7B-Inst & NQ+HQA    & 38.6 & 34.6 & 16.2 & 29.8 \\
\midrule
\multicolumn{7}{l}{\emph{Different setup (non-comparable backbone or training mix)}} \\
R1-Searcher$^{\ddagger}$ & 7B-\textbf{Base} & \textbf{HQA+2Wiki} & 44.2 & 51.3 & 15.8 & 37.1 \\
StepSearch$^{\dagger}$   & 7B-Inst & MuSiQue & 34.8 & 36.3 & 13.4 & 28.2 \\
GlobalRAG$^{\dagger}$    & 7B-Inst & 2Wiki+HQA & 38.2 & 47.8 & 14.9 & 33.6 \\
\midrule
\multicolumn{7}{l}{\emph{Our reward ablations: same setup as Search-R1 (Qwen2.5-7B-Inst, NQ+HQA)}} \\
\quad Search-Only          & 7B-Inst & NQ+HQA & 38.6 & 34.6 & 16.2 & 29.8 \\
\quad Binary Gap           & 7B-Inst & NQ+HQA & 36.6 & 39.0 & 24.6 & 33.4 \\
\quad Typed Base           & 7B-Inst & NQ+HQA & 37.8 & 40.5 & 25.5 & 34.6 \\
\quad \textbf{Typed+Shape (ours)} & 7B-Inst & NQ+HQA
                           & \textbf{39.1} & \textbf{41.5}
                           & \textbf{26.7} & \textbf{35.4} \\
\bottomrule
\end{tabular}
\\[2pt]
\raggedright\footnotesize
$^{\dagger}$ Reproduction numbers as reported in
\cite{globalrag2025} on the standard Search-R1 dev splits.
$^{\ddagger}$ R1-Searcher reports Cover-EM in its own paper; the strict-EM
numbers shown here are an independent reproduction by \cite{ragr12025}.
R1-Searcher trains on HotpotQA+2Wiki, so HotpotQA and 2Wiki are
in-domain for it; for our NQ+HotpotQA mix, 2Wiki and MuSiQue are
out-of-domain.
\end{table*}

\textbf{Reward ablation.}
We note that 2Wiki and MuSiQue are out-of-domain w.r.t.
our NQ$+$HotpotQA training mix while HotpotQA is
in-domain; the consistent Typed+Shape gain on both
suggests the typed signal generalizes beyond the
training distribution. 

Concretely, in the single-run macro-average setting,
Typed+Shape improves EM by $+2.4$ over Binary Gap
(typed labels matter) and by $+1.2$ over Typed Base
(behavioral shaping matters). Without shaping, the
generator can detect a gap yet still commit to an answer;
with shaping, mean search count rises from $2.5$ to $2.8$,
consistent with the policy learning to re-search after
IE/MB detections. The three-seed Qwen setting (Appendix~\ref{app:grpo_robustness},  Table~\ref{tab:grpo_robustness}) confirms the same monotone ordering, with Typed+Shape improving over Search-Only by $+3.3$ EM.
On strictly out-of-domain MuSiQue (neither system trained on it), Typed+Shape outperforms R1-Searcher by $+10.9$~EM
($26.7$ vs.\ $15.8$); the latter's higher Avg comes from its HotpotQA+2Wiki in-domain training data rather than a stronger reward design.

\begin{figure}[t]
\centering
\includegraphics[width=\columnwidth]{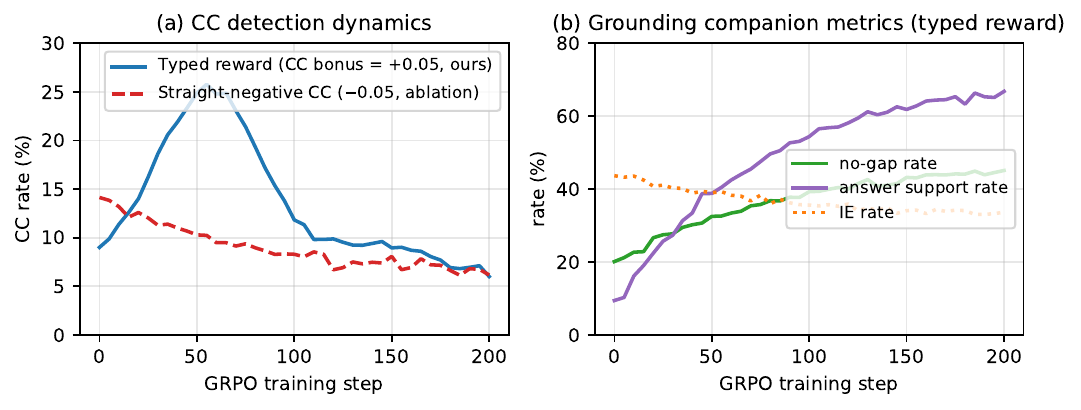}
\caption{Two-phase CC dynamics during 200 GRPO iterations. \textbf{(a)} CC rate under our typed reward (blue, +0.05 bonus) vs.\ a straight-negative CC ablation (red dashed). \textbf{(b)} Companion grounding metrics (no-gap rate, answer support rate, IE rate) under the typed reward.}
\label{fig:cc_dynamics}
\end{figure}

\textbf{Two-phase CC dynamics.}
The CC detection bonus produces a characteristic non-monotone
rate during training (Figure~\ref{fig:cc_dynamics}): CC rises
early as the generator surfaces contradictions for the bonus,
then falls as it learns to avoid making contradicted claims
altogether. A straight-negative CC reward decreases CC
monotonically but achieves it through vaguer, harder-to-check
claims that hurt support rate; the positive bonus prevents
this contradiction-suppression failure mode.

\textbf{Llama backbone transfer.}
\label{sec:grpo_robustness}
On Llama-3.1-8B under identical hyperparameters, all checker-augmented rewards underperform Search-Only ($28.1$): Binary Gap collapses to $19.2$, Typed+Shape reaches $25.4$. We attribute this to backbone-specific interaction with our Qwen-tuned reward magnitudes (see Limitations) and treat Qwen as the headline; the Untyped-Dense control is in Appendix~\ref{app:exp_grpo}.



\section{Conclusion}

We presented \textbf{StepGap}, a hybrid NLI-LLM checker that emits typed step-level evidence-gap labels (\textsc{CC}/\textsc{IE}/\textsc{MB}), each anchored at a specific NLI decision and tied to a concrete repair action. Used as a typed GRPO process reward, StepGap improves Qwen2.5-7B-Instruct EM by $+3.3$ across three seeds. Typed+Shape obtains  $71.5\%$ answer-claim support, the highest among our reward ablations; a distilled Qwen2.5-7B student preserves $87\%$ of the teacher's step-F1 without API calls at inference. Methodologically, we hope to leave the reader with the view that two checkers within each other's F1 CIs can still expose qualitatively different process signals---one through monotone components, the other through competing-error cancellation---and only the former is usable as a typed RL reward.

\section*{Limitations}
\paragraph{Sample size and statistical power.}
Our human-annotated  evaluation set covers 181 steps across 82
questions; bootstrap $95\%$ CIs on sF1 overlap between
StepGap, LLM-Only, and several baselines
(Table~\ref{tab:variants_sf1}). We therefore frame our
contribution on the architectural and diagnostic axes
(\emph{healthy ablation pattern}, \emph{Q-F1 trap},
\emph{typed signal}) rather than as a top-line F1 win,
and we report seed-level variance in Appendix~\ref{app:grpo_robustness}  (Table~\ref{tab:grpo_robustness}). As partial mitigation of
the small gold sample, we additionally report
StepGap-vs-teacher agreement on the $1{,}000$-step
distillation pool (\S\ref{sec:lens1_sf1}, binary
$\mathrm{F1}{=}0.88$, typed $\mathrm{F1}{=}0.81$),
showing that StepGap's binary
gap-detection behavior is consistent at $\sim 5\times$ the
gold-benchmark scale. Low-frequency gap types (\textsc{MB}, $n{=}9$) remain the main power-limited case, and a larger MB-targeted gold evaluation set is left to future work. We note, however, that our 181-step scale is comparable to step-level annotation in the agentic-RAG literature, and the silver-standard $1{,}000$-step check addresses the architectural-level concern at $\sim 5\times$ scale.

\paragraph{Backbone-specific reward tuning.}
The reward magnitudes (\S\ref{sec:reward}) were tuned on Qwen2.5-7B and do not transfer zero-shot to Llama-3.1-8B (Table~\ref{tab:grpo_robustness}); we hypothesize interaction between dense per-step shaping and Llama's instruction-following distribution under a fixed KL budget, and leave per-backbone retuning to future work.
\paragraph{Other limitations.}
The shaping term relies on surface pattern matching for
retraction detection (e.g., ``actually'', ``wait''),
which may miss subtler rewordings and, conversely, is
in principle vulnerable to reward hacking via spurious
retraction-pattern usage. We did not observe such hacking
in our $200$-iterations runs, but longer training could enable
this failure mode and motivates a learned retraction
classifier in future work.

\bibliography{custom}

\appendix
\section{GRPO Robustness: Seeds, Backbones, Untyped Control}
\label{app:grpo_robustness}

Table~\ref{tab:grpo_robustness} reports GRPO EM
(mean$\pm$std over 3 seeds) on Qwen2.5-7B-Instruct and
Llama-3.1-8B-Instruct under identical hyperparameters.
On Qwen, the ordering Search-Only $<$ Binary Gap $<$
Typed Base $<$ Typed+Shape is monotone with all
seed-level std $\le 1.2$ EM, well below the $+3.3$
headline gain. On Llama all checker-augmented rewards
underperform Search-Only; we discuss this backbone-specific failure in \S\ref{sec:grpo_robustness} and
explicitly limit our generalization claim to the Qwen
backbone in the Limitations section. The
\emph{Untyped-Dense} control row (matched per-step
density to Typed+Shape but no CC/IE/MB distinction)
isolates the typing-vs-density confound on Qwen; the
sparser Binary Gap row in Table~\ref{tab:rl_results}
provides a complementary density-matched control.

\begin{table}[h]
\centering
\small
\caption{GRPO EM (mean$\pm$std over 3 seeds) across reward
designs and two generator backbones. On Qwen2.5-7B,
Typed+Shape gives a clear ordering
Search-Only $<$ Typed Base $<$ Typed+Shape; on
Llama-3.1-8B, all checker-augmented rewards
\emph{underperform} Search-Only, with Binary Gap
collapsing most severely. We discuss this backbone-
specific failure in \S\ref{sec:grpo_robustness} and treat
the Qwen result as the primary headline result.}
\label{tab:grpo_robustness}
\begin{tabular}{lcc}
\toprule
\textbf{Reward} & \textbf{Qwen2.5-7B} & \textbf{Llama-3.1-8B} \\
\midrule
Search-Only        & 32.1 $\pm$ 0.3 & \textbf{28.1} $\pm$ 0.7 \\
Binary Gap         & 31.0 $\pm$ 0.4 & 19.2 $\pm$ 0.5 \\
Typed Base         & 34.2 $\pm$ 1.2 & 28.3 $\pm$ 0.7 \\
Typed+Shape (ours) & \textbf{35.4} $\pm$ 0.9 & 25.4 $\pm$ 0.4 \\
\bottomrule
\end{tabular}
\end{table}

\section{Annotation Protocol}
\label{app:annotation}

\paragraph{Annotators.} Two annotators with graduate-level NLP
training labeled the test subset of 82 questions / 181 steps.
Both were familiar with multi-hop QA conventions but were
naive to the StepGap taxonomy at the start of training.

\paragraph{Training.} Annotators completed a 30-minute
calibration session on a held-out set of 20 steps, with
discussion of edge cases (in particular, the boundary between
\textsc{Missing Bridge} and \textsc{Irrelevant Evidence}).
Decision rules were finalized as: a step is IE if the
retrieved evidence is about a different entity than the
step's claim addresses; MB if the evidence is about the right
entity but does not entail the claim; CC if the evidence
contradicts the claim or the step targets the wrong relation.

\paragraph{Disagreement resolution.} Steps with disagreement
were adjudicated by a third senior annotator. Final labels
were the post-adjudication consensus.

\paragraph{Per-type agreement.} Cohen's $\kappa$ on the binary
\texttt{has\_gap} decision is 0.704. Per-type $\kappa$ values
are: IE 0.71, CC 0.68, MB 0.43. The lower $\kappa$ on MB is
driven by its low frequency ($n{=}9$ true MB cases) and
inflated sampling variance, not by systematic disagreement.

\paragraph{Annotation interface.} Annotators saw the question,
the full reasoning trace, the retrieved evidence per step,
and the final answer. They were asked to assign one of
\{no-gap, CC, IE, MB\} per step plus a brief free-form
justification.

\paragraph{Licenses and artifact use.}
HotpotQA, 2WikiMultiHopQA, and MuSiQue are released under their
original research licenses (CC BY-SA 4.0, Apache 2.0, and CC BY 4.0
respectively); we use them only for academic evaluation, consistent
with the original intended use. The Wikipedia 2018 dump used for
retrieval follows the Wikipedia text license (CC BY-SA 3.0).
E5-base-v2 and DeBERTa-v3 are released under MIT; Qwen2.5-7B-Instruct
under the Tongyi Qianwen license; Llama-3.1-8B under the Llama 3.1
Community License. We release the StepGap checker code, the
distilled student checkpoint, and the 181-step evaluation benchmark
under CC BY 4.0 for non-commercial research use.

\section{Knowledge Distillation to Local Student}
\label{app:distillation}

\paragraph{Setup.}
We distill the hybrid teacher (GPT-4.1-mini + DeBERTa-v3-NLI)
into a fully local Qwen2.5-7B student via supervised
fine-tuning on 1,000 teacher-labeled steps (4 datasets excluding
the test benchmark). Training uses LoRA ($r=16$) on a single
H100 GPU; total cost is 2 hours + \$8 teacher API.

\paragraph{Results.}
Table~\ref{tab:distill} compares teacher, student, and an
ablated student without NLI input features.

\begin{table}[h]
\centering
\small
\caption{Distillation results. F1 and $\kappa$ vs.\ the human-annotated set.}
\label{tab:distill}
\begin{tabular}{lccc}
\toprule
\textbf{Checker} & \textbf{F1} & \textbf{$\kappa$} & \textbf{Cost/200 steps} \\
\midrule
Teacher (4.1-mini+NLI) & .720 & .284 & \$0.06 \\
Student 7B (v4)        & .626 & .264 & \$0 \\
Student 7B (no NLI)    & .582 & .198 & \$0 \\
Student 8B (v5)        & .603 & .060 & \$0 \\
\bottomrule
\end{tabular}
\end{table}

We report two student variants. The Qwen3-8B v5 row
exhibits near-zero $\kappa$ ($.060$), which we attribute
to thinking-mode KV-cache behavior at our LoRA rank
($r{=}16$) producing unstable Stage-C outputs; we therefore
adopt Qwen2.5-7B v4 as the primary student. 

The 7B student retains $87\%$ of teacher F1 \emph{and} removes teacher API calls at inference time,
making the checker suitable for integration into long RL
runs without per-step API expense. Cohen's $\kappa$ drops
from $.284$ (teacher) to $.264$ (student), a small
absolute decline of $0.02$. We note that both values are
low in absolute terms; this reflects the inherent difficulty of typed step-level annotation -- already evident in the human $\kappa{=}0.704$ on $n{=}181$ -- rather than student degradation. The retained signal is sufficient for
downstream typed reward: substituting the student for the
teacher in the GRPO loop yields $+1.7$ EM over Search-Only
(vs. $+3.3$ with the teacher,
Table~\ref{tab:rl_results}) -- about half the teacher's
gain at zero API cost, which we view as a favorable
cost-quality trade-off for long RL runs.

\paragraph{StepGap-vs-teacher agreement on the
distillation pool.}
The $1{,}000$-step distillation pool also serves as a
silver-standard validation set for StepGap itself. We
measure agreement between StepGap (the deployed system,
GPT-4.1-mini Stages~A--C + DeBERTa-v3 Stages~D--E) and a
held-out teacher run on the same $1{,}000$ steps,
reporting binary $\mathrm{F1}{=}0.88$ on
\texttt{has\_gap} and typed $\mathrm{F1}{=}0.81$ on
CC/IE/MB. The gap between the silver typed F1
($0.81$) and student typed F1 ($.626$,
Table~\ref{tab:distill}) localizes student degradation
to capacity rather than supervision noise: the teacher
signal that the student is being trained on is itself
internally consistent at scale.
\paragraph{Student failure analysis.}
Manual inspection of 47 student false negatives reveals one
dominant pattern: \textbf{\texttt{found\_quote $\to$ inference
$\to$ no-gap} over-triggering}. The student reports
\texttt{found\_quote=True} at Stage C even when the ``quote''
is from the wrong entity or does not address the claim. At
this point the pipeline short-circuits to no-gap,
bypassing the NLI entailment check that the teacher performs.
23 of 47 FN cases (49\%) match this pattern. The bottleneck
is not model capacity (the 7B student matches teacher F1 on
explicit-gap cases) but a \emph{calibration bias} in the
student's Stage C judgment. Hard-negative augmentation on $26$ teacher-confirmed FN cases raises student F1 by $3.7$ points (Appendix~\ref{app:exp_hardneg}).
\section{Hybrid Architecture Rationale}
\label{app:hybrid_rationale}

\paragraph{Why hybrid, not LLM-only.}
Step-level gap detection requires several distinct judgments:
entity-relation alignment, abstention validity, entity
consistency, verbatim-quote extraction, logical entailment,
and cross-step verification. Conflating these into one LLM
prompt invites \emph{prompt context contamination}: framing
introduced by an early node biases later judgments. The
Whitehorse case in Appendix~\ref{sec:case_studies} shows this
in isolation -- Stage~A's ``airline vs.\ terminal'' framing
pushes a shared-context Stage~D away from accepting an
otherwise entailing quote. The hybrid breaks this by making
Stage~D and Stage~E pure NLI calls that receive only the
\texttt{(premise, hypothesis)} pair.

\paragraph{Hybrid ablation pattern.}
Table~\ref{tab:ablation_hybrid_vs_llm_only} reports parallel
component-removal ablations on the LLM-Only and StepGap
checkers. The two architectures show qualitatively
different ablation profiles:
\emph{(i) LLM-Only shows competing-error cancellation}: three of four component
removals \emph{improve} F1 ($+1.6$, $+0.8$, $+1.7$),
indicating internal components actively cancel each other's
errors. Only one component (Global Prior) is a net positive
contributor.
\emph{(ii) StepGap is healthy}: every component removal hurts
F1 ($-6.6$ for Stage~A, $-2.5$ for Stage~E
cross-verification), indicating each stage is a positive
contributor to the final decision. This compositional
property is the architectural signal that motivates StepGap
over LLM-Only at equal top-line F1.

\paragraph{Why only Stages A and E.} 
Stage~B (abstention validity) fires on too few steps in our 
evaluation set ($n{<}10$) to support a meaningful 
component-wise ablation. Stages~C (entity consistency) 
and~D (local entailment) form a tightly coupled 
LLM$\to$NLI pipeline: Stage~D consumes Stage~C's verbatim 
quote as its NLI premise, so removing either alone 
collapses the decision tree to a different topology 
rather than yielding a clean ablation. We therefore 
ablate the two stages with structurally independent input 
pathways---Stage~A (pre-retrieval alignment) and Stage~E 
(cross-step verification)---both of which show 
$-\mathrm{F1}$, confirming the monotone pattern.
\begin{table}[h]
\centering
\small
\caption{Component-removal ablations on the 181-step evaluation set.
\textbf{LLM-Only} (top): three of four removals improve F1, a
competing-error cancellation pattern. \textbf{StepGap} (bottom):
every removal hurts F1, the healthy pattern that motivates the
hybrid design. Common-Knowledge / Semantic-Leap / Global  Prior denote the LLM-Only checker's internal sub-prompts  for general-knowledge inference, paraphrase tolerance,  and prior probability of gap, respectively.}
\label{tab:ablation_hybrid_vs_llm_only}
\begin{tabular}{lcccc}
\toprule
\textbf{Configuration} & \textbf{P} & \textbf{R}
                       & \textbf{F1} & \textbf{$\Delta$F1} \\
\midrule
\multicolumn{5}{l}{\emph{LLM-Only checker}} \\
Full system            & 69.2 & 71.2 & 70.0 & --- \\
\; $-$ Stage~A (alignment)
                       & 73.0 & 70.2 & 71.6 & \textbf{+1.6} \\
\; $-$ Common-Knowledge
                       & 70.5 & 71.2 & 70.8 & \textbf{+0.8} \\
\; $-$ Semantic-Leap
                       & 75.5 & 68.3 & 71.7 & \textbf{+1.7} \\
\; $-$ Global Prior
                       & 69.3 & 58.7 & 63.5 & $-$6.5 \\
\midrule
\multicolumn{5}{l}{\emph{StepGap (ours)}} \\
Full system            & 71.0 & 73.1 & 72.0 & --- \\
\; $-$ Stage~A (alignment)
                       & 69.3 & 58.7 & 63.5 & $-$6.6 \\
\; $-$ Stage~E (cross-verify)
                       & 67.0 & 68.3 & 67.6 & $-$2.5 \\
\bottomrule
\end{tabular}
\end{table}
\paragraph{Auditability.}
Each decision is unit-testable. NLI outputs are deterministic
given model weights, and the label mapping
(\textsc{entailment}~$\to$~no-gap,
\textsc{neutral}~$\to$~MB,
\textsc{contradiction}~$\to$~CC) is closed-form. Debugging a
misclassification reduces to isolating quote extraction
(Stage~C, LLM) from entailment (Stage~D, NLI). An LLM-only
checker offers no such isolation.

\paragraph{Distillability and reward grounding.}
The hybrid is a prerequisite for two downstream uses. (i) Local
deployment: the NLI sub-system is already a 435M model, and
Stage~A--C produce structured JSON that a 7B student can
imitate (Appendix~\ref{app:distillation}). (ii) Typed reward:
the CC/IE/MB labels are grounded in the deterministic NLI
output, which an LLM-only checker cannot match because its
categorical judgments shift across reruns and prompt
perturbations.

\section{Failure Mode Taxonomy of the Full System}
\label{app:fm_taxonomy}

Inspection of 13 checker false negatives and 12 false
positives reveals four recurrent failure modes. We describe
them using the typed taxonomy.

\paragraph{FM1: Parametric shortcut (FN source).}
Generator retrieves partially relevant evidence; NLI assigns
entailment because a surface token overlap exists; generator
fills in the remaining fact from parametric memory, producing
a wrong answer. No CC/IE/MB is flagged.

\paragraph{FM2: Granularity mismatch (FN source, 6/13).}
Generator retrieves a related but lower-granularity fact (year
instead of birth date, district instead of municipality);
NLI sees the entity at some granularity and judges entailment.
The answer is wrong but the step passes. This is the most
common FN.

\paragraph{FM3: Comparison logic (FP source).}
Conclusion steps for comparison questions (``who died later?'')
require combining evidence from multiple prior steps; NLI
judges each step against its direct evidence pool in isolation,
often producing \textsc{neutral} and spurious MB flags.

\paragraph{FM4: Answer string normalization (FP source).}
Generator produces a correct answer with different surface form
from the gold string (``La Vida Alrededor was released earlier''
vs.\ gold ``La Vida Alrededor''); the question counts as wrong
under EM, but the checker correctly detects no evidence gap.
This is a limitation of EM-based ground truth, not of the
checker.

\section{Case Studies}
\label{sec:case_studies}

We illustrate how StepGap handles each gap type with two representative examples drawn from our human-annotated evaluation set. Two further cases (a student-checker failure mode and a Single-LLM-XL failure mode) are deferred to Appendix~\ref{app:cases}.

\paragraph{Case 1: NLI breaks prompt contamination
(Whitehorse, 2WikiMultiHopQA).}
\emph{Question}: ``The airline operating in Whitehorse, Yukon
handled how many passengers in 2012?''
\emph{Retrieved evidence}: ``The terminal handled
$294{,}000$ passengers in 2012.''
\emph{Gold}: $294{,}000$.

\begin{table}[h]
\centering
\small
\renewcommand{\arraystretch}{1.15}
\begin{tabular}{p{0.21\columnwidth}p{0.31\columnwidth}p{0.31\columnwidth}}
\toprule
\textbf{Stage} & \textbf{LLM-Only}
               & \textbf{StepGap (ours)} \\
\midrule
A: alignment   & on-target $\to$ pass
               & on-target $\to$ pass \\
C: quote search
               & finds ``terminal handled $294{,}000$ passengers''
               & finds ``terminal handled $294{,}000$ passengers'' \\
D: entailment  & LLM judges \textsc{contradiction}
                 (``terminal $\neq$ airline'')
               & NLI judges \textsc{entailment} \\
\midrule
\textbf{Result} & \textbf{CC (false positive)}
                & \textbf{no-gap (correct)} \\
\bottomrule
\end{tabular}
\end{table}

\noindent The two systems run on identical inputs but
disagree at Stage~D. LLM-Only's Stage~A introduced an
``airline vs.\ terminal'' framing into the shared prompt,
which biased its Stage~D judgment away from accepting the
quote. StepGap's NLI Stage~D receives only the
\texttt{(premise, hypothesis)} pair, with no upstream framing,
and judges entailment correctly. Air~North is the sole
commercial airline operating from the Whitehorse terminal,
so the quote does support the claim. This is the
context-contamination mechanism we discuss in
Appendix~\ref{app:hybrid_rationale}, isolated to a single
decision.
\paragraph{Case 2: Missing-Bridge detection
(Kuhn vs.\ Pertramer, 2WikiMultiHopQA).}
\emph{Question}: ``Were Heinrich Gerhard Kuhn and Elfie
Pertramer from the same country?''
\emph{Gold}: yes (both German).
\emph{Predicted answer}: no (the model concluded
``not from same country'').

\begin{table}[h]
\centering
\small
\renewcommand{\arraystretch}{1.15}
\begin{tabular}{p{0.18\columnwidth}p{0.65\columnwidth}}
\toprule
\textbf{Step} & \textbf{StepGap decision} \\
\midrule
1. ``Kuhn is a British physicist''
& Stage~D: NLI=\textsc{entailment} (quote found, supported)
$\to$ no-gap \\
2. ``Pertramer is a German actress''
& Stage~D: NLI=\textsc{entailment} $\to$ no-gap \\
3. ``Britain $\neq$ Germany''
& common-knowledge inference, no quote needed
$\to$ no-gap \\
\textbf{4. ``Therefore not from same country''}
& \textbf{Stage~D: NLI=\textsc{neutral}; quote found but does
not entail (``British physicist'' specifies nationality, not
country of birth) $\to$ \textsc{MB}} \\
\bottomrule
\end{tabular}
\end{table}

\noindent StepGap correctly identifies the conclusion step as
\textsc{Missing Bridge}: the retrieved evidence supports the
sub-claim ``Kuhn is British'' but \emph{not} the unstated
bridging premise ``Kuhn was born in Britain.'' The actual
bridge is that ``British physicist'' refers to academic
affiliation, not birthplace; Kuhn was born in Germany. This
is the kind of unstated-premise failure that endpoint EM
cannot catch (the answer is wrong, but no claim is overtly
contradicted) and that a binary support-checker would also
miss (the quote is supported, just incomplete). The repair
action prescribed by an MB tag (``issue a bridging search'')
is exactly what the generator would need to do to recover.

\section{Additional Case Studies}
\label{app:cases}

We supplement the two case studies in
\S\ref{sec:case_studies} with two further examples that
illustrate failure modes of alternative checkers, motivating
specific design choices in StepGap.

\paragraph{Case 3: Student over-triggers \texttt{found\_quote}
(Peter and Paul Fortress, MuSiQue).}
\emph{Question}: ``Where did the designer of Peter and Paul
Fortress die?''
\emph{Gold}: Saint Petersburg.
\emph{Evidence}: ``Peter and Paul Fortress is the original
citadel of St.\ Petersburg, Russia, built to Domenico
Trezzini's designs.''

\begin{table}[h]
\centering
\small
\renewcommand{\arraystretch}{1.15}
\begin{tabular}{p{0.21\columnwidth}p{0.31\columnwidth}p{0.31\columnwidth}}
\toprule
\textbf{Stage} & \textbf{Teacher (GPT-4.1-mini + NLI)}
               & \textbf{Student (Qwen2.5-7B distilled)} \\
\midrule
C: quote
               & finds ``citadel of St.\ Petersburg''
               & also finds it; sets \texttt{found\_quote=True} \\
D: entailment  & NLI=\textsc{neutral} (citadel location $\neq$
                 designer's place of death)
               & \emph{(short-circuited; never invoked)} \\
\midrule
\textbf{Result} & \textbf{MB (correct)}
                & \textbf{no-gap (false negative)} \\
\bottomrule
\end{tabular}
\end{table}

\noindent The student's failure is structural: it learns
during distillation to map ``\texttt{found\_quote=True}
$\wedge$ inference step'' directly to no-gap, bypassing the
NLI entailment check the teacher performs. This pattern
accounts for 23 of 47 (49\%) student false negatives
(\S\ref{app:distillation}). The fix is not more capacity
but \emph{stronger Stage~C calibration}: the student must
learn that ``found a quote'' is necessary but not sufficient
for no-gap. We address this via hard-negative augmentation
(briefly described in Appendix~\ref{app:distillation}).

\paragraph{Case 4: Single-LLM-XL flags everything
(HotpotQA).}
\emph{Question}: ``Which 2017 film was directed by the
director of \emph{Whiplash}?''
\emph{Gold}: \emph{La La Land}.
\emph{Evidence at step 1}: ``\emph{Whiplash} is a 2014 film
directed by Damien Chazelle.''

\begin{table}[h]
\centering
\small
\renewcommand{\arraystretch}{1.15}
\begin{tabular}{p{0.21\columnwidth}p{0.31\columnwidth}p{0.31\columnwidth}}
\toprule
\textbf{Stage} & \textbf{StepGap (ours)}
               & \textbf{Single-LLM-XL} \\
\midrule
A: alignment   & on-target
               & on-target \\
C: quote       & finds ``directed by Damien Chazelle''
               & ``no exact-match quote'' (over-strict) \\
D/E            & NLI=\textsc{entailment} $\to$ no-gap
               & --- (skipped) \\
\midrule
\textbf{Result} & \textbf{no-gap (correct)}
                & \textbf{IE (false positive)} \\
\bottomrule
\end{tabular}
\end{table}

\noindent A single LLM, even a strong one, applies its quote
search and entailment criteria too uniformly, treating any
trace of paraphrasing or compositional inference as
unsupported. Across the 181-step evaluation set this collapses
into the 87.8\% IE rate reported in
\S\ref{sec:lens2_dist}---and produces the same 91.4\% Q-F1
that we identify in \S\ref{sec:lens1_sf1} as the Q-F1
trap. The decomposed pipeline in StepGap, by contrast,
accepts paraphrastic matches at Stage~C and only enforces
strict entailment at Stage~D, where the inputs are
proper propositions.

\section{Reward Hyperparameters}
\label{app:hparams}

\paragraph{Base reward weights.} The base reward magnitudes
$\{+0.20, -0.05, -0.10, +0.05\}$ for $\{$no-gap, MB, IE,
CC$\}$ were selected on a 200-iteration pilot, optimizing
answer-claim support rate at fixed EM. The relative
ordering (no-gap $\gg$ CC, MB $>$ IE) was held fixed; only
the absolute scale was tuned. For the CC detection bonus
specifically, we tested
$\{+0.02, +0.05, +0.10\}$ at single seed: $+0.02$ produced
the same monotone CC-suppression as the negative ablation
(Figure~\ref{fig:cc_dynamics}), $+0.10$ caused CC rate to
plateau without the second-phase decline, and $+0.05$
exhibited the two-phase rise-then-fall dynamics we report
as the intended design behavior.

\paragraph{Shaping weights.} The shaping reward magnitudes
$\{+0.10, +0.15, -0.15, -0.05\}$ were chosen so that the
total per-step reward magnitude does not exceed the EM reward
magnitude (which is 1.0 binary).

\paragraph{Near-duplicate query threshold.}
The token-F1 threshold $0.7$ used in $r^{\text{shape}}$ to 
detect near-duplicate queries (\S\ref{sec:shape_reward}) was 
set on the same 200-iteration pilot as the base reward 
weights: lower values incorrectly flagged paraphrastic 
reformulations as duplicates and dampened re-search behavior, 
while higher values failed to flag near-identical retries 
differing only in stop words or token order. We leave a 
fine-grained sweep of this threshold to future work.

\paragraph{Lambda.} The reward mixing coefficient $\lambda$
defaults to 1.0. Sensitivity to $\lambda \in \{0.5, 1.0, 2.0\}$
shows EM varies within $\pm 0.6$ points; we report
$\lambda = 1.0$ as the main configuration.

\paragraph{Limitations.} A learned reward weighting (e.g.,
inverse-propensity per gap type) is left to future work.

\section{Step-Category Distribution}
\label{app:lens2_dist}

Table~\ref{tab:cat_dist} reports the per-variant
distribution of assigned gap types over the 181-step
evaluation set. A well-calibrated checker should produce IE in
the $38$--$41\%$ range (the rate observed on LLM-Only and
StepGap); deviations from this band reflect systemic biases
that sF1 alone may not penalize.

\begin{table}[h]
\centering
\small
\caption{Step-category distribution across checker variants
on the 181-step evaluation set. Healthy IE band: $38$--$41\%$.
$\dagger$~marks degenerate variants whose IE rate exceeds
$70\%$ with collapsed CC and MB.}
\label{tab:cat_dist}
\begin{tabular}{lcccc}
\toprule
\textbf{Variant} & \textbf{no-gap} & \textbf{IE}
                 & \textbf{CC}     & \textbf{MB} \\
\midrule
LLM-Strict             & 27.6 & 40.9 & 17.1 & 14.4 \\
LLM-Only               & 40.9 & 39.2 & 14.9 &  5.0 \\
\textbf{StepGap (ours)} & \textbf{40.9} & \textbf{38.1}
                       & \textbf{14.9} & \textbf{6.1} \\
\midrule
Single-LLM-XL$^\dagger$ & 12.2 & 87.8 &  0.0 &  0.0 \\
NER-Quote$^\dagger$     & 14.4 & 74.0 & 10.5 &  1.1 \\
\bottomrule
\end{tabular}
\end{table}

The two pathologies sF1 alone cannot make visible:
\emph{Single-LLM-XL} assigns zero steps to CC and zero to MB,
reducing the three-way taxonomy to a binary (no-gap vs.\ IE)
decision -- a stronger single LLM produces a checker that
\emph{cannot discriminate between gap types}. This is precisely
the property that disqualifies it from typed-reward use:
typed reward needs all three categories reliably emitted.
\emph{NER-Quote} replaces Stage~C's LLM quote search with
spaCy NER plus token overlap; the result is $74\%$ IE because
Stage~D's NLI call fires on noun-phrase queries that are not
well-formed propositions, so the checker over-reports
retrieval failure. The hybrid works because LLM and NLI
handle \emph{different factor types}, not because NLI is
smaller or because an LLM is larger.

\section{Answer-Gap Correlation}
\label{app:gap_correlation}

Table~\ref{tab:gap_correlation} cross-tabulates final-answer
correctness with gap detection on the 82-question subset.
Of 69 wrong-answer questions, $56$ ($81\%$) have at least one
gap detected. Of $13$ correct-answer questions, $12$ still
have a gap detected -- the \emph{lucky-correct} regime: a
substantial fraction of correct answers in multi-hop QA are
not fully grounded in retrieved evidence.

\begin{table}[h]
\centering
\small
\caption{Answer correctness vs.\ gap detection ($n=82$
questions).}
\label{tab:gap_correlation}
\begin{tabular}{lcc}
\toprule
 & \textbf{Gap Detected} & \textbf{No Gap Detected} \\
\midrule
Wrong Answer   & 56 (TP) & 13 (FN) \\
Correct Answer & 12 (FP) &  1 (TN) \\
\bottomrule
\end{tabular}
\end{table}

\section{First-Gap-Type Localization}
\label{app:first_gap_localization}

For each wrong-answer question we examine the first step the
checker flags as a gap and record its type
(Table~\ref{tab:first_gap}). StepGap and LLM-Only both
identify IE (retrieval failure) as the first broken step in
roughly $80\%$ of wrong-answer chains, confirming that
retrieval, not reasoning, is the dominant root cause of
failure in our multi-hop evaluation set. StepGap differs from
LLM-Only in striking a more balanced first-gap distribution
between IE ($79\%$), MB ($11\%$), and CC ($11\%$), giving
downstream repair logic a richer typed signal. Degenerate
variants attribute essentially every first failure to IE,
producing no usable signal for routing repair actions.

\begin{table}[h]
\centering
\small
\caption{For each wrong-answer question, type of the first
step flagged by the checker as containing a gap. Entries are
percentages of wrong-answer questions. Degenerate variants
trivially attribute nearly all first-failures to IE.}
\label{tab:first_gap}
\begin{tabular}{lccc}
\toprule
\textbf{Variant} & \textbf{First-IE} & \textbf{First-CC}
                 & \textbf{First-MB} \\
\midrule
LLM-Strict               & 62  & 12 & 26 \\
LLM-Only                 & 80  & 12 &  7 \\
\textbf{StepGap (ours)}  & 79  & 11 & 11 \\
\midrule
Single-LLM-XL & 100 (deg.) & --- & --- \\
NER-Quote     &  97 (deg.) & --- & --- \\
\bottomrule
\end{tabular}
\end{table}

\section{Q-F1 Trap: Empirical Confirmation}
\label{app:qf1_trap_empirical}

We complement the closed-form argument in
Section~\ref{sec:lens1_sf1} with two empirical checks
(Table~\ref{tab:qf1_trap_empirical}). \emph{First}, on four
public QA datasets at their natural wrong-answer rate $w$
(estimated from a Qwen2.5-7B-Instruct Search-Only baseline), a
flag-everything dummy checker reaches a Q-F1 within $\pm 0.02$
of the analytic prediction $2w/(1+w)$. \emph{Second}, on
stratified resamples of our own evaluation set with $w$ controlled
at $0.50$, $0.65$, and $0.84$, empirical Q-F1 reproduces the
analytic curve, confirming that the effect is driven by $w$
and the metric, not by the specific question content.
\begin{table}[t]
\centering
\small
\setlength{\tabcolsep}{4pt} 
\caption{Empirical Q-F1 of a flag-everything checker
on public benchmarks (top) and stratified resamples
of our evaluation set (bottom). Analytic prediction:
$2w/(1+w)$.}
\label{tab:qf1_trap_empirical}
\begin{tabular}{p{0.30\columnwidth}ccc}
\toprule
\textbf{Dataset / stratum} & $w$
  & \textbf{Q-F1} & \textbf{(analytic)} \\
\midrule
\multicolumn{4}{l}{\emph{Cross-dataset, natural $w$}} \\
NQ            & 0.55 & 0.70 & (0.71) \\
HotpotQA      & 0.68 & 0.80 & (0.81) \\
2Wiki         & 0.76 & 0.85 & (0.86) \\
MuSiQue       & 0.90 & 0.94 & (0.95) \\
\midrule
\multicolumn{4}{l}{\emph{Stratified resamples, $w$ controlled}} \\
Stratum-50    & 0.50 & 0.66 & (0.67) \\
Stratum-65    & 0.65 & 0.79 & (0.78) \\
Stratum-84    & 0.84 & 0.91 & (0.91) \\
\bottomrule
\end{tabular}
\end{table}

\section{GRPO Auxiliary Grounding Metrics}
\label{app:rl_grounding}

Table~\ref{tab:rl_grounding} reports the auxiliary grounding
metrics that accompany the EM numbers in
Table~\ref{tab:rl_results}: token-F1, answer support rate
(fraction of answer claims NLI-entailed by a retrieved
quote), and mean search calls per question. Across reward
configurations, support rate rises monotonically
($0\% \to 38.2\% \to 56.1\% \to 71.5\%$), and search count
increases under typed shaping ($2.1 \to 2.8$), indicating
that the policy learns to issue additional searches in
response to detected gaps rather than answering through.

\begin{table}[h]
\centering
\small
\caption{Auxiliary grounding metrics for each reward
ablation on the 82-Q subset (aggregate over the three
datasets). \emph{Supp.}~$=$ fraction of answer claims
NLI-entailed by a retrieved quote;
$|\texttt{<s>}|$~$=$ mean search calls per question.}
\label{tab:rl_grounding}
\begin{tabular}{lccc}
\toprule
\textbf{Configuration} & \textbf{F1} & \textbf{Supp.}
                       & \textbf{$|\texttt{<s>}|$} \\
\midrule
Search-Only        & 39.6          & ---           & 2.1 \\
Binary Gap         & 40.4          & 38.2          & 2.3 \\
Typed Base         & 41.6          & 56.1          & 2.5 \\
Typed+Shape (ours) & \textbf{43.1} & \textbf{71.5} & \textbf{2.8} \\
\midrule
w/ Student checker & 41.2          & 48.3          & 2.6 \\
w/ Teacher checker & 43.1          & 71.5          & 2.8 \\
\bottomrule
\end{tabular}
\end{table}

\section{Threshold-Sweep Diagnostic}
\label{app:thresh_sweep}

To verify that the degenerate classification of Single-LLM-XL and
NER-Quote (\S\ref{sec:lens2_dist}) is not threshold-dependent,
we sweep confidence thresholds for these variants and the
StepGap reference. Per-variant confidence is the geometric mean
of stage-level confidences: for LLM stages, the softmax
probability of the chosen output token; for NLI stages, the
softmax probability of the chosen entailment label.

\begin{figure}[t]
\centering
\includegraphics[width=\columnwidth]{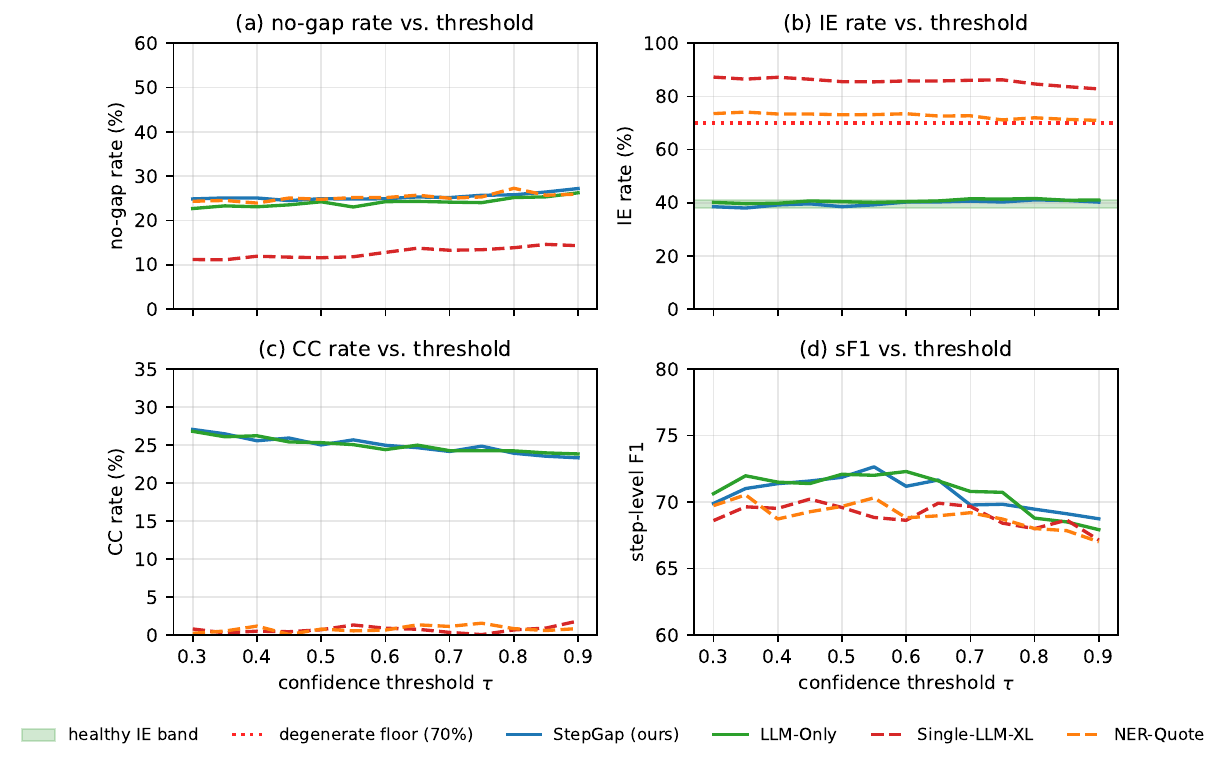}
\caption{Threshold sweep for StepGap (healthy reference) vs.\
Single-LLM-XL and NER-Quote (degenerate variants). Top-right panel:
green band marks the healthy IE range (38--41\%) observed
across LLM-Only and StepGap; red dashed line marks the
70\% degenerate floor. Single-LLM-XL and NER-Quote never enter
the healthy band at any threshold $\tau \in [0.3, 0.9]$.}
\label{fig:threshold_sweep}
\end{figure}

\section{Checker Implementation Prototype}
\label{app:code}

This appendix provides a minimal reference implementation of
the StepGap hybrid checker. The full production code handles
batching, caching, retrieval-server integration, and robust
JSON parsing; here we present the core decision logic in
stripped-down form. The five-stage pipeline
(Figure~\ref{fig:pipeline}) is implemented as a linear
sequence of early-return checks.

\subsection{Structured-Output Schema}

The LLM (GPT-4.1-mini in the teacher; Qwen2.5-7B in the
student) returns a single structured JSON object per step,
containing the judgments for Stage~A, Stage~B, and Stage~C in
one pass. Structured-output mode is enforced via
\texttt{response\_format=\{"type":"json\_schema",\dots\}} with
\texttt{strict=true}, which guarantees parseable JSON. The
schema is:

{\small\begin{verbatim}
{
  "step_minus1_alignment": {
    "is_off_target":       bool,
    "drift_type":  // one of:
      "none" | "entity_drift" |
      "relation_drift" | "scope_drift",
    "alignment_reasoning": str
  },
  "step_0_abstention_check": {
    "is_abstention_step":     bool,
    "abstention_is_accurate": bool,
    "abstention_reasoning":   str
  },
  "step_1_quote_search": {
    "entity_match":           bool,
    "entity_match_reasoning": str,
    "found_quote":            bool,
    "evidence_quote":         str,
    "quote_search_reasoning": str
  }
}
\end{verbatim}}

\subsection{LLM Prompt Templates}
\label{app:prompts}

Stages~A--C are invoked via a single prompt that returns the
schema above. The prompt has a fixed three-section structure
(alignment / abstention / entity-and-quote) and switches
between an \emph{inference} variant and a \emph{conclusion}
variant of Section~A based on the step type. We reproduce the
core text below; whitespace and formatting follow the actual
production prompt.

\paragraph{Section A: alignment (inference variant).}
{\footnotesize\begin{verbatim}
STEP -1: SUBQUESTION ALIGNMENT
Does this step target the CORRECT entity and
relation?
CRITICAL: retrieval failure != off-target.
Only flag if step_text ITSELF names the wrong
entity.

EC quick test - is_off_target = FALSE if:
 - Entity in step_text appears verbatim in
   the question
 - Entity is the correct intermediate result
 - step_text names correct film/person but
   retrieval returned wrong doc

TRUE drift: Q asks "AUTHOR of book" -> step
  searches "DIRECTOR of film adaptation"
NOT drift: retrieval failed but query is
  correct

If is_off_target = TRUE -> has_gap = TRUE,
gap_type = "unsupported_claim", STOP.
\end{verbatim}}

\paragraph{Section A: alignment (conclusion variant).}
{\footnotesize\begin{verbatim}
STEP -1: CONCLUSION ALIGNMENT (mandatory)
RULE 0: answer contains N/A or unknown
  -> is_off_target = FALSE, bypass to Node 0.

Otherwise C-1 -> C-2 -> C-3:
  C-1: target attribute type
       (PLACE / BIRTH_YEAR / SPOUSE /
        AGENT / COUNTRY / YES_NO)
  C-2: actual type of answer value
  C-3: TYPE mismatch  = relation_drift.
       Wrong value but same type
                       = NOT drift.

DRIFT:    Q "Where born?" -> "Naresh Kumar"
          (person != place) -> TRUE
NO DRIFT: Q "Who did X?"  -> "Viktor Belash"
          (person  = person) -> FALSE
\end{verbatim}}

\paragraph{Section B: abstention.}
{\footnotesize\begin{verbatim}
STEP 0: ABSTENTION CHECK
(skip if Step -1 triggered)
Is this step saying N/A / cannot determine?
  Evidence lacks it
    -> GROUNDED ABSTENTION
       is_abstention_step = True,
       abstention_is_accurate = True
  Evidence has it
    -> WRONG ABSTENTION
       is_abstention_step = True,
       abstention_is_accurate = False
\end{verbatim}}

For conclusion steps with prior inference evidence available,
the prompt prepends a \texttt{GLOBAL EVIDENCE} block listing
the snippets retrieved at all earlier steps; Stage~B uses this
pool to decide whether an asserted abstention is justified.

\paragraph{Section C: entity consistency + verbatim quote.}
{\footnotesize\begin{verbatim}
STEP 1: ENTITY CONSISTENCY + QUOTE SEARCH
(skip if Step -1 or Step 0 triggered)

Part A - Entity consistency check:
  EC-1: What entity/topic does this step's
        query target?
  EC-2: What entity/topic is the retrieved
        evidence actually about?
        (check the document title and first
         sentence)
  EC-3: Same entity? -> entity_match = TRUE,
                        proceed to quote search
        Different    -> entity_match = FALSE,
                        found_quote  = FALSE
                        (stop here)

  *Quote from the WRONG document does NOT
   count.*
   Step:     "Sruthilayalu composer"
   Evidence: "M.G. Sreekumar" (different
              person) -> FALSE

Part B - Verbatim quote search
        (only if entity_match = TRUE):
  Find a 5-20 word EXACT span from evidence.
  Must be: (i) exact text,
           (ii) about the same entity,
           (iii) relevant to the step claim.
\end{verbatim}}

The prompt closes with the question, the previous-step
summaries (truncated to the last three), the current step's
text, and the evidence pool serialized as JSON.

\paragraph{NLI calls (Stage~D and Stage~E).}
The NLI stages take no prompt: they pass a
\texttt{(premise, hypothesis)} pair directly to a
DeBERTa-v3-NLI cross-encoder \cite{laurer2024deberta}. Stage~D
sets \texttt{premise}~$=$~the verbatim quote returned by
Stage~C and \texttt{hypothesis}~$=$~the step's claim text;
Stage~E iterates over entity-matched snippets from prior steps
as the premise pool, with the same hypothesis. We softmax the
three-class logits and take the argmax label
(\textsc{entailment} / \textsc{neutral} /
\textsc{contradiction}); ties default to \textsc{neutral}.

\subsection{NLI Label Calibration}

NLI models from the \texttt{cross-encoder} family emit a
three-dimensional logit vector whose label order varies by
checkpoint (e.g.\
\texttt{[contradiction, entailment, neutral]} or
\texttt{[entailment, neutral, contradiction]}). At checker
initialization we probe with a known entailment pair
(``A cat is an animal.'', ``A cat is an animal.'') and
identify which output dimension scores highest; the label
order is set accordingly. All subsequent calls use the
calibrated mapping. Thresholds are \texttt{ENTAILMENT=0.5} and
\texttt{CONTRADICTION=0.5} on the softmaxed scores; labels
between these thresholds are treated as \textsc{neutral}.

\subsection{Step Type Classification}

A step is classified as \texttt{inference} if its text
contains a \texttt{<search>} tag (or, in Hermes-style outputs,
a \texttt{<tool\_call>} JSON object with the \texttt{search}
action), and as \texttt{conclusion} if it contains an
\texttt{<answer>} tag. Ambiguous steps (neither tag) are
treated as \texttt{inference} by default.

\subsection{Pipeline Path Accounting}

Every check returns a \texttt{pipeline\_path} string encoding
the exact sequence of stage decisions taken (e.g.\
\texttt{stageC:entity\_mismatch},
\texttt{stageD:neutral}). We use this for two purposes:
(i) distributional analysis, confirming each stage is actually
exercised in practice and not starved of input, and identifying
which branches dominate the error modes; (ii) unit testing,
where we assert that specific annotated cases follow the
expected path. The pipeline-path signature also underlies our
distillation loss: the student is trained to reproduce the
teacher's full path, not just the terminal label, which
prevents the student from short-circuiting via surface
heuristics (see Section~\ref{app:distillation}).

\section{Additional Experimental Details}
\label{app:experiment_specs}

This appendix documents experimental protocols and
reproducibility details that complement the main paper.

\subsection{Standard-Split Re-evaluation Protocol}
\label{app:exp_stdsplits}

For comparability with the published baselines in
Table~\ref{tab:rl_results} (top panel), we follow the
standard Search-R1 evaluation pipeline on the standard dev
splits (sizes in Table~\ref{tab:benchmark_stats}):
Qwen2.5-7B-Instruct generator with temperature $0$ and
max-new-tokens $1024$; E5-base-v2 retrieval
\cite{wang2022e5} over Wikipedia 2018 with top-$k{=}5$;
GRPO-trained checkpoint at the best validation EM step,
chosen on a held-out $500$-Q dev pool disjoint from these
splits. Each cell reports strict EM averaged over the full
split. Our reward-ablation rows in
Table~\ref{tab:rl_results} use the same pipeline.

\subsection{GRPO Training Configuration}
\label{app:exp_grpo}

All GRPO runs hold the following hyperparameters constant:
group size $G{=}8$, KL coefficient $\beta{=}0.05$, learning
rate $1{\times}10^{-6}$ for the policy with the checker
frozen, $3{,}000$ training iterations with $200$-iterations
evaluations, on the NQ$+$HotpotQA training mix specified in
Section~\ref{sec:lens4_downstream}. The
\emph{Untyped-Dense} control reward referenced in
Section~\ref{sec:grpo_robustness} uses
$r^{\text{base}}$ that maps any gap (CC, IE, MB) to
$-0.10$ and no-gap to $+0.20$, with no shaping term and no
per-type distinction; this isolates the contribution of
\emph{typing} from per-step density.

\paragraph{Computing infrastructure.}
All GRPO training is run on a single NVIDIA H100 (80GB) GPU.
A full 3,000-iteration GRPO run takes approximately 18 H100~GPU-hours
for Qwen2.5-7B-Instruct and 22 H100~GPU-hours for Llama-3.1-8B-Instruct
under our setup. Teacher-checker API usage (GPT-4.1-mini) is
approximately 72K calls per run at \$22; distillation of the student
checker costs \$8 in teacher API plus 2 H100~GPU-hours. The 181-step
human evaluation set was processed once per checker variant, totalling
under 0.5 H100~GPU-hours of additional inference. Total compute across
all reported configurations (three seeds on Qwen, single seed on Llama,
all reward ablations, distillation, and Appendix experiments) is
approximately \textbf{ 350-400} H100~GPU-hours plus.
\subsection{Hard-Negative Augmentation Protocol}
\label{app:exp_hardneg}

The $+3.7$~F1 gain reported for the student in
Appendix~\ref{app:distillation} comes from augmenting the
student's training data with $26$ hard-negative cases
identified by manual inspection of teacher--student
disagreement on the held-out distillation pool. A case is
selected if the teacher's NLI label and the student's NLI
label diverge \emph{and} the teacher's prediction matches
the human label on the 181-step evaluation set for an adjacent
step from the same trace. Augmented training uses the same
LoRA configuration ($r{=}16$) for $200$ additional iterations
with a learning rate of $5{\times}10^{-7}$ (an order of
magnitude smaller than initial distillation, so as not to
overfit the small augmentation set). We hold out $13$ of
the $26$ hard-negative examples as a within-augmentation
check to confirm the gain is not attributable to
memorization.

\end{document}